%% file: acl_latex.tex
\pdfoutput=1

\documentclass[11pt]{article}

\usepackage[final]{acl}

\usepackage{times}
\usepackage{latexsym}

\usepackage[T1]{fontenc}

\usepackage[utf8]{inputenc}

\usepackage{microtype}

\usepackage{inconsolata}

\usepackage{graphicx}

\usepackage{authblk}

\usepackage{lipsum}
\usepackage{booktabs}
\usepackage{array}
\usepackage{pifont} 
\newcommand{\cmark}{\ding{51}} 

\usepackage{float}
\usepackage{subcaption}
\usepackage{multirow}

\title{Finding A Voice: Exploring the Potential of \\ African American Dialect and Voice Generation for Chatbots}

\author[1]{\bf Sarah E. Finch}
\author[2]{\bf Ellie S. Paek}
\author[1]{\bf Ikseon Choi}
\author[2]{\bf Jinho D. Choi}

\affil[1]{School of Nursing, Emory University, Atlanta, GA, USA}
\affil[2]{Department of Computer Science, Emory University, Atlanta, GA, USA}

\affil[ ]{\texttt{\{sfillwo, ellie.paek, ike.choi, jinho.choi\}@emory.edu}}

\begin{document}
\maketitle

\input{latex/sections/0_abstract}
\input{latex/sections/1_introduction}

\input{latex/sections/2_related-work}
\input{latex/sections/3_approaches}
\input{latex/sections/4_experiments}
\input{latex/sections/6_discussion}

\input{latex/sections/7_conclusion}

\clearpage
\input{latex/sections/limitations}
\input{latex/sections/10_acknowledgements}

\bibliography{custom}

\clearpage
\input{latex/sections/8_appendix}

\end{document}

%% file: latex/sections/0_abstract.tex
\begin{abstract}

As chatbots become integral to daily life, personalizing systems is key for fostering trust, engagement, and inclusivity. This study examines how linguistic similarity affects chatbot performance, focusing on integrating African American English (AAE) into virtual agents to better serve the African American community. We develop text-based and spoken chatbots using large language models and text-to-speech technology, then evaluate them with AAE speakers against standard English chatbots. Our results show that while text-based AAE chatbots often underperform, spoken chatbots benefit from an African American voice and AAE elements, improving performance and preference. These findings underscore the complexities of linguistic personalization and the dynamics between text and speech modalities, highlighting technological limitations that affect chatbots' AA speech generation and pointing to promising future research directions.

\end{abstract}

%% file: latex/sections/1_introduction.tex
\section{Introduction}

In recent years, chatbots have become increasingly popular, assisting users in a variety of tasks across various applications \cite{alsharhan_survey:24}. As these systems become more embedded in daily life, personalized human-computer interaction has become crucial. Personalizing chatbots can enhance their effectiveness by tailoring interactions to individual preferences \cite{zhang_personalization:24, huang_personalization:24}. Research in human interactions shows that matching professionals with clients based on shared characteristics, such as ethnicity, can improve rapport and trust \cite{street_hh_similarity:08, takeshita_hh_similarity:20}. These findings have inspired efforts to personalize chatbots based on interpersonal similarity as well.

Studies on chatbot personalization through interpersonal similarity have explored directions of both visual and linguistic aspects, but results are mixed. Research on visual similarity suggests that aligning a chatbot's avatar skin tone with that of the user can boost satisfaction and engagement \cite{liao_similarity:20, park_similarity:24}. However, studies on linguistic similarity have shown varied outcomes, with some reporting benefits \cite{agarwal_codeswitching_hinglish_21} and others less favorable results \cite{obremski_cultural_representation_speech:22}. As of now, consistent approaches for effective personalization through linguistic similarity remain elusive.

This study examines the impact of linguistic similarity on human-chatbot interactions to better understand effective personalization. While previous research has focused on multilingual contexts \cite{arora_codeswitching_comix_23, liu_codeswitching_education:24}, the nuances of dialectal variations within a single language are less explored. It has been suggested that integrating dialects into chatbot design could improve identity alignment and trust \cite{martin_speaking:24}. Our research focuses on African American English (AAE), a key dialect extensively used by the African American community \cite{rickford_aave:99}. This group faces unique challenges in technology adoption due to the historical stigmatization of AAE and its underrepresentation in natural language processing tasks \cite{blodgett_twitter:18,koenecke_asr:20,ziems_value:22}. This lack of representation perpetuates the view that chatbots cannot effectively process AAE, potentially discouraging African American users from engaging with these technologies \cite{harrington_codeswitching:22}. We aim to explore the benefits of incorporating AAE into chatbot responses to enhance personalization and acceptance.

Our research seeks to advance chatbots proficient in AAE through a twofold approach. First, we develop text chatbots using Large Language Models (LLMs) that generate AAE responses. We evaluate three LLM families for their effectiveness as AAE text chatbots, varying in AAE feature expression. Second, we convert these text chatbots into speech using a text-to-speech model that produces a voice with an African American accent. Both text and spoken AAE chatbots are evaluated by AAE speakers on key performance characteristics, comparing them to Standard American English (SAE) chatbots. Our findings highlight the critical role of chatbot modalities in integrating AAE. While AAE text chatbots do not perform well, spoken chatbots with an African American voice and subtle AAE features are favored by African American users. This performance contrast underscores the complexity of chatbot personalization and our analysis offers valuable insights for future advancements.\footnote{Our code and data is publicly released at \url{https://github.com/emorynlp/AAVE-Chat}.}

%% file: latex/sections/2_related-work.tex
\begin{figure*}[t]
    \centering
\includegraphics[width=\textwidth]{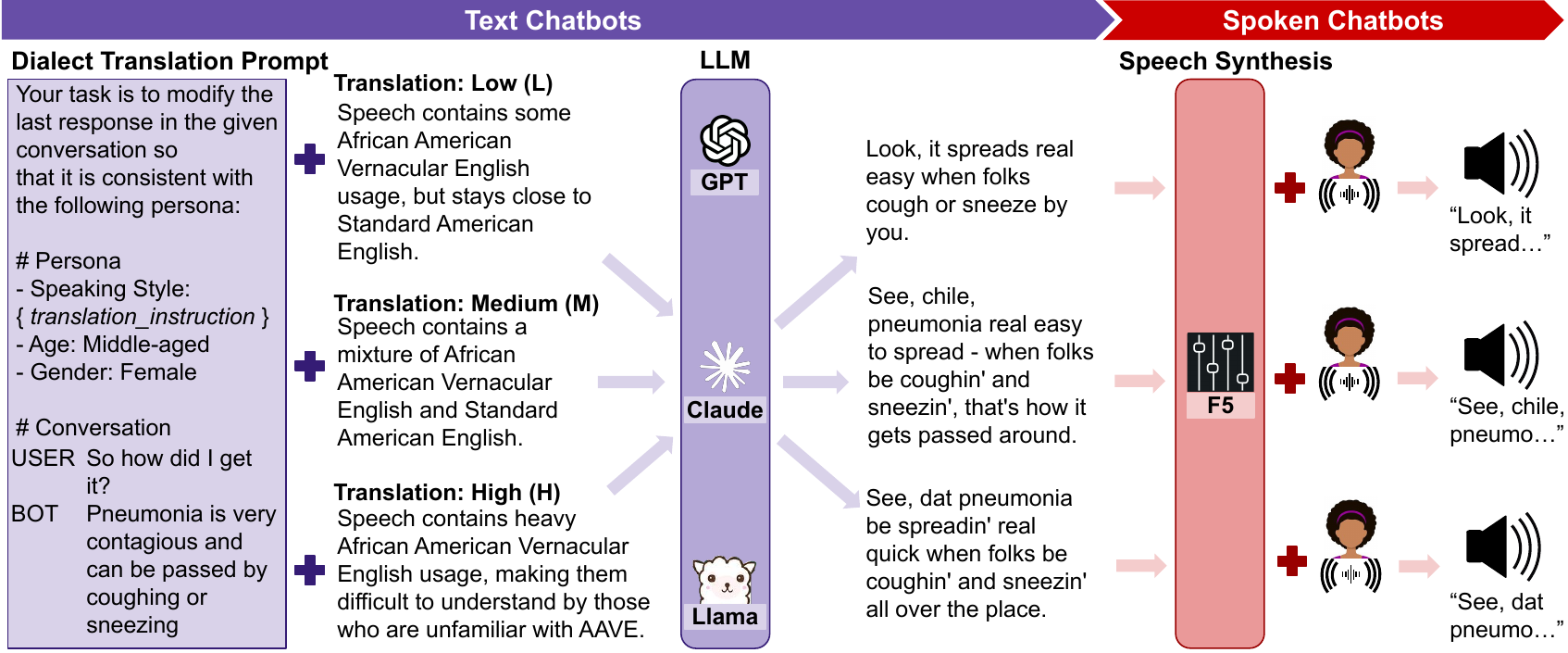}
    \caption{Overview of the dialect translation and voice generation approaches taken for Text and Spoken Chatbots.}
    \label{fig:chatbots-approach}
    \vspace{1ex}
\end{figure*}

\section{Related Work} 


\paragraph{AAE Dialect}

African American English (AAE), also referred to as African American Vernacular English (AAVE) or African American Language (AAL), represents a systematic variety of English with distinct phonological, morphosyntactic, and lexical features that have been extensively documented in sociolinguistic literature \cite{rickford_aave:99, green_aave:02, sidnell_aaetest:02, wolfram_aaetest:04}. Key features include the habitual "be" construction ("She be working late") that indicates recurring actions, the completive "done" marker ("I done finished my homework") signaling completed actions with emphasis, the absence of third person present tense marking ("He walk to school"), multiple negation which intensifies rather than cancels negation ("I don't know nothing about that"), and final consonant cluster reduction ("han'" instead of "hand") \cite{rickford_aave:99}. We leave detailed definitions of AAE and its features to previous literature. In this work, we draw on such established literature to inform the development and evaluation of AAE dialect features in African American (AA) chatbots, while leveraging input from real-world AAE speakers to assess chatbot performance through realistic usage experiences.

\paragraph{Modeling Culture}

Recent research has explored how Large Language Models (LLMs) can tailor their responses based on specific cultural contexts, which have achieved success especially within question-answering applications \cite{jin_personalization_ip:24, putri_cultural_representation_commonsense:24}. Additionally, there is increasing concern over stereotypes being reinforced and negative impacts on minority groups, finding that LLM outputs can be biased based on user racial cues \cite{wan_bias:23, kantharuban_bias:24, fleisig_bias:24}. These findings highlight the need to carefully consider personalization in LLMs to avoid negative effects, thus motivating our research into the impact of AAE usage by chatbots.

\paragraph{Language Accommodation}
Research into language accommodation in chatbots has focused on their ability to perform code-switching, or the blending of multiple languages in a single conversation, with studies showing that chatbots capable of this linguistic flexibility tend to show greater empathy compared to their monolingual counterparts \cite{bhattacharya_impact_codeswitching_empathy:24} and have improved task performance and social rapport when engaging with users who rely on similar multilingual capacities \cite{brixey_impact_codeswitching_diaagent:24, choi_impact_codeswitching_multiingual:23} 
While the existing research centers on code-switching between widely spoken languages, such as English-Spanish or English-Hindi \cite{arora_codeswitching_comix_23, agarwal_codeswitching_hinglish_21, liu_codeswitching_education:24}, incorporating minority dialects into chatbot systems is thought to be equally valuable \cite{martin_speaking:24}. In line with this perspective, our study specifically investigates the integration of the English dialect of AAE into chatbot outputs.

\paragraph{Generation of AAE}

There has been ongoing research to develop models that translate Standard American English (SAE) into African American English (AAE). These approaches include training models on specially constructed datasets \cite{groenwold_aavegen:20, graves_aavegen:24}, utilizing Large Language Models (LLMs) \cite{deas_bias:23}, and creating syntax-rule-based systems derived from linguistic feature analyses \cite{ziems_multivalue:23}. The systems based on LLMs deliver promising results, generating AAE outputs that are both natural and accurate, albeit still falling short of the quality achieved with SAE generation \cite{deas_bias:23}. 
A limitation of prior research is their reliance on tweet datasets \cite{groenwold_aavegen:20, deas_bias:23}, which differ significantly in structure and context from chatbot dialogues \cite{blodgett_twitter:18}, or query-response interactions, rather than exploring the complexities of multi-turn dialogues \cite{fleisig_bias:24}. Moreover, much of the research has exclusively emphasized bias analysis, often overlooking broader aspects of user experience with AAE-enabled virtual assistants \cite{wan_bias:23, fleisig_bias:24}. Critically, there is a lack of systematic studies controlling the level of AAE used in chatbot interactions, which hinders the assessment of how variations in dialect intensity might affect chatbot performance. We address these limitations by conducting experiments that systematically vary AAE intensity and evaluating the impact on multiple aspects of chatbot performance in multi-turn dialogues.

\paragraph{Generation of Accented Speech}

Previous research has explored the development of accented voices through voice cloning and deep learning models \cite{vinotha_accents_speech_synthesis:24, nechaev_accent_conversion:24}. While many studies concentrate on accents from different countries, some focus on regional dialects within a country \cite{falai_dialect_tts:22, pazylbekov_dialect_learning:19}. Despite advancements, research is limited on how accent-integrated chatbots impact users who share these accents. Some studies examine how accents influence user impressions, but often rely on feedback from the general population and lack disclosure of participants' ethnic backgrounds \cite{piercy_accent_sterotypes:25, jones_tutor_accent:24}. Research specific to accent-integrated chatbots within their intended demographic shows negative impressions of such dialogue agents but focus on accent adaptation between English and non-English languages, rather than dialectal variations \cite{obremski_cultural_representation_speech:22}. A notable study recently trained an African American voice model using a voice actor \cite{pinhanez_aa_tts:24}, though the model was not publicly released and was not tested in the context of chatbot technology. In this work, we contribute to the advancement of accented chatbots through evaluating the performance of AA-accented chatbots for AAE speakers.

%% file: latex/sections/3_approaches.tex
\section{African American Chatbot Design}

The development of chatbots that share linguistic similarities with the African American community involves two key aspects. Firstly, there is the distinctive African American English (AAE) dialect, known for its unique linguistic characteristics that set it apart from other English variants.\footnote{Not all African Americans speak AAE, nor are all AAE speakers African American, though an estimated 80\% of African Americans do \cite{rickford_aave:99}.} These features span phonological, morphological, syntactical, and semantic dimensions \cite{rickford_aave:99}. Secondly, research indicates that African American speakers also exhibit a unique accent with its own particular tone and prosody \cite{pinhanez_aa_tts:24}. Our approach to chatbot development addresses both these aspects (Fig. \ref{fig:chatbots-approach}). For Text Chatbots, we focus on incorporating the AAE dialect into written chatbot responses (Sec. \ref{sec:text-chatbots}). For Spoken Chatbots, we incorporate an African American accent into spoken chatbot responses (Sec. \ref{sec:spoken-chatbots}).

\subsection{Text Chatbots: AAE Dialect}
\label{sec:text-chatbots}

For the generation of chatbot responses in the AAE dialect, we choose to model dialect expression separately from response generation. Namely, we treat dialect expression as a function $E(I, D_a, D_b) \rightarrow O$ that translates a string input $I$ in dialect $D_a$ into a string output $O$ in dialect $D_b$. By distinguishing between response generation and dialect expression, we minimize the risk of demographic-related biases inherent to specific dialects influencing the content of responses \cite{fleisig_bias:24}. This separation ensures that the effect of dialect is confined to surface-level stylistic elements, leaving the response's semantics unaffected.

In this study, we implement the function $E$ as a call to an LLM using an SAE-to-AAE translation prompt, such that $D_a = SAE$ and $D_b = AAE$. We construct 3 prompt variants, where each variant adjusts the strength of the instruction to express the response in AAE such that three levels of AAE expression are obtained: Low, Medium, and High. Figure \ref{fig:chatbots-approach} illustrates the three translation variants.

We include 3 popular LLM families for study: Llama3.1-70b \cite{dubey_llama:24}, Claude-sonnet-3.5 \cite{anthropic_claude:24}, and GPT4o \cite{hurst_gpt:24}. In total, we test 9 different Text Chatbots, where each chatbot is a different combination of translation prompt and LLM.\footnote{Full prompts, chatbot configurations, and example outputs are provided in Appendices \ref{app:prompts}, \ref{app:model-hyperparameters}, and \ref{app:dialect-output-examples}.}


\subsection{Spoken Chatbots: AAE Dialect \& Accent}
\label{sec:spoken-chatbots}

To develop spoken AAE chatbots, we employ a text-to-speech (TTS) model to convert the text responses into speech. Specifically, we use the F5 model proposed by \citet{chen_f5:24}, a high-performing, publicly available TTS system. F5 is a non-auto-regressive approach based on the Diffusion Transformer \cite{peebles_diffusiontransformer:23} and ConvNeXt V2 \cite{woo_convnext:23}, trained on text-guided speech-infilling \cite{le_voicebox_infill:24}. This model generates speech conditioned on both input text and a speaker audio reference, enabling accent production alongside AAE-specific linguistic patterns.
To generate an African American accented voice, we extract a short audio clip from an interview in the publicly available Corpus of Regional African American Language \cite{kendall_coraal:23}, selecting a speaker matching the persona used for AAE dialect generation (Fig. \ref{fig:chatbots-approach}) based on demographic data provided in the corpus. To represent the human user in dialogues, we use a short audio clip of a Standard American voice from LibriSpeech \cite{panayotov_librispeech:15}.\footnote{Since the user's voice is not the focus of this study, we use an SA voice to distinguish chatbot speech from user speech.} We include 4 Spoken Chatbots for study, each representing one combination of AAE dialect level (None/SAE, Low, Medium, or High) and AA voice accent. We use the AAE responses from the best-performing Text Chatbot, which is detailed in Section \ref{sec:aa-spoken-results}.

Each dialogue utterance is independently transformed into speech, utilizing the corresponding speaker reference. To enhance audio quality, we first preprocess the utterance text based on manual testing. This involves converting symbols such as numbers, dollar signs, and percentages into words, and dividing lengthy utterances into smaller segments using the spaCy sentence splitter \cite{honnibal_spacy:20}. Once preprocessed, each segment is individually converted into speech. All audio segments for each dialogue are then concatenated to create complete dialogue files, with brief pauses inserted between speaker turns to facilitate natural-sounding turn-taking. Additionally, a visual cue displaying the speaker's name (whether it's the user or the chatbot) accompanies the audio for each speaker in the final dialogue video file, in order to easily identify who is currently speaking.






%% file: latex/sections/4_experiments.tex
\section{Experiments}

To measure the capabilities of the AA Text Chatbots and AA Spoken Chatbots, we perform two evaluations, measuring AAE dialect feature expressions (Section \ref{sec:aae-feature-experiment}) and AA chatbot performance (Section \ref{sec:aae-chatbot-experiment}). 

\subsection{Data}
\label{sec:chatbot-applications}

We utilize multi-turn dialogues from the extensive LLM-generated dialogue dataset, SODA \cite{kim_soda:23}, as evaluation data. This dataset is particularly valuable as it includes speaker labels that help categorize interactions by role, such as "Doctor" for Healthcare and "Teacher" for Education. By leveraging these labels, we can selectively extract dialogues that align with popular chatbot applications. Based on chatbot surveys, we identify 5 popular chatbot applications: Customer Assistance, Commerce, Healthcare, Education, and Social Companionship \cite{alsharhan_survey:24,motger_survey:22,luo_survey:22,caldarini_survey:22,rapp_survey:21,chaves_survey:21,adamopoulou_survey:20}. Using appropriate speaker labels,\footnote{Table \ref{tab:popular-chatbot-domains} in Appendix \ref{app:popular-chatbot-applications} indicates the full list of roles used.} we obtain subsets of dialogues per domain and sample 20 10-turn dialogues per domain to create a comprehensive set of 100 dialogues suitable for our evaluation. The turns corresponding to the domain role are treated as chatbot turns in each dialogue, and are converted to AAE using the approach in Section \ref{sec:text-chatbots} for AA Text Chatbots and then converted to audio using the approach in Section \ref{sec:spoken-chatbots} for AA Spoken Chatbots.

We choose static generation over live chatbot interactions to maintain experimental control and isolate the specific effects of dialect variation. By using pre-existing dialogue datasets to generate AA chatbot outputs, we can make direct comparisons across models without the confounding variables that interactive evaluations would introduce due to variability in dialogue content and flow. This static evaluation approach ensures that all chatbots are assessed on identical content and aligns with standard methodological practices in dialogue research.

\subsection{Baselines}

To comprehensively discern the influence of dialect and accent on chatbot performance, we include baseline chatbots that utilize the SAE dialect and SA accent for comparison. For the Text Chatbots, the baseline is a chatbot using the SAE dialect. This is accomplished through constructing a translation instruction for the same prompt as that used for AAE response generation that instructs the LLM to output the response in the SAE dialect. For Spoken Chatbots, the baseline is a chatbot using the Standard American accent. This is accomplished by using an additional short audio clip from the LibriSpeech dataset of a Standard American speaker.

\subsection{AAE Dialect Feature Expression}
\label{sec:aae-feature-experiment}

We first quantify and validate the usage of various AAE features in the responses from the AA Text Chatbots in order to analyze their behavior. Namely, we want to measure the rate of phonetic, morphological, syntactical, and semantic changes that the Text Chatbots make to the dialogue responses when tasked with translating them to AAE. In order to do this, we develop an automatic approach for tagging AAE linguistic features present in the generated responses, which leverages a large language model to identify and label spans in the response that incorporate AAE features.

\begin{figure}[t]
    \centering
    \includegraphics[width=.9\columnwidth]{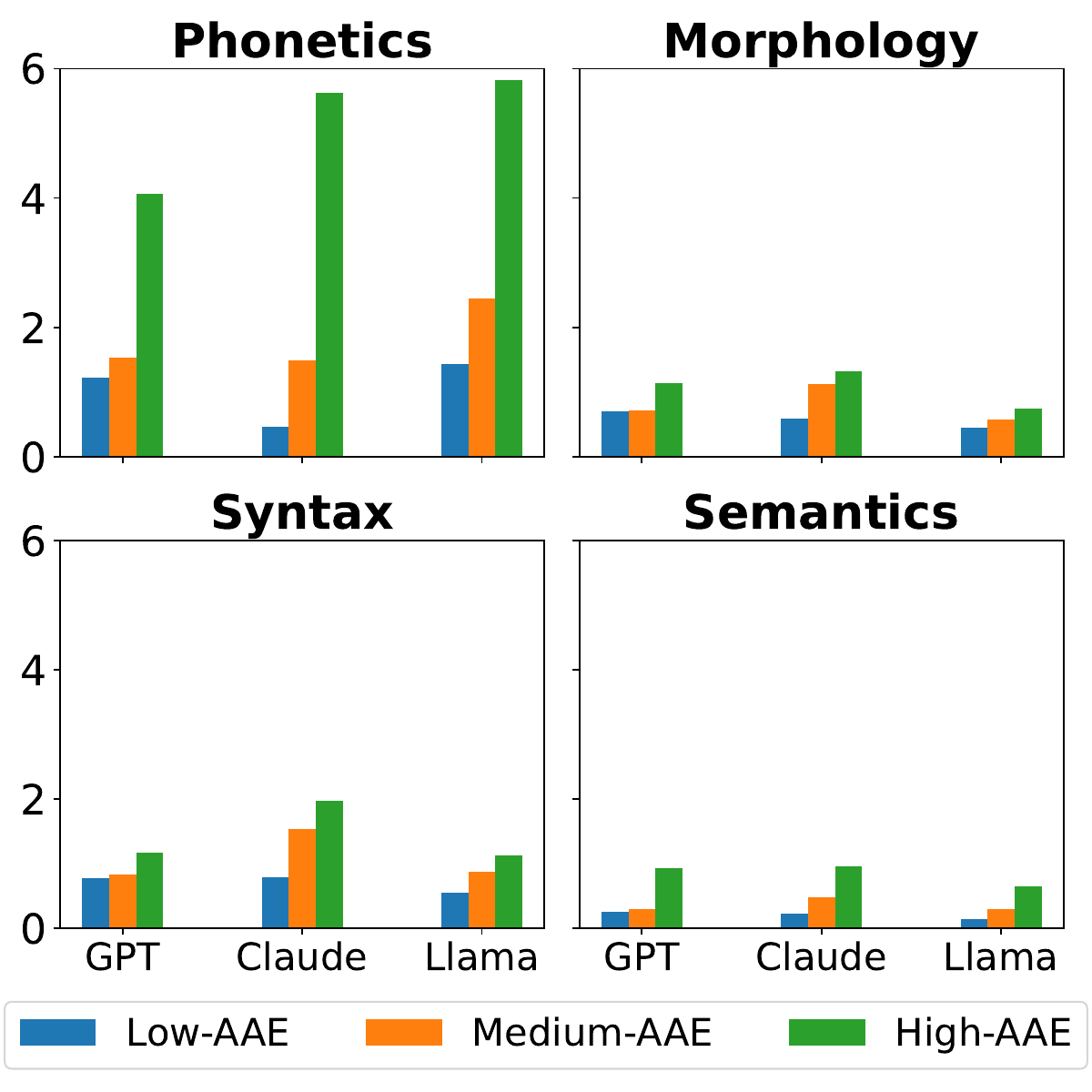}
    \caption{Comparison of average per-turn rates of AAE features in Text Chatbot responses.}
    \label{fig:aave-output-analysis}
    \vspace{-2ex}
\end{figure}

To ensure the accuracy of this tagging approach, we create a test set comprising AAE text alongside extracted spans labeled with their AAE linguistic features. This test set is constructed using labeled examples from existing AAE literature and resources.\footnote{Examples of AAE test cases are provided in Appendix \ref{app:aae-tagging-examples}.} Overall, the AAE feature test data consists of 90 texts containing a total of 136 feature labels, covering over 30 of the most common AAE features. We conduct experiments using both GPT-4o and Claude-Sonnet-3.5 for feature tagging, finding that Claude-Sonnet-3.5 outperforms GPT-4o with an accuracy of 91\% compared to 86\% in feature identification. Using Claude-Sonnet-3.5, we apply feature tagging to half ($n=250$) of the translated responses generated by the 9 AA Text Chatbots under study. Figure \ref{fig:aave-output-analysis} displays the distribution of AAE features across each Text Chatbot.

Phonetic features emerge as the most prominent AAE feature generated by the Text Chatbots, particularly within the High dialect chatbots, which average over 3 phonetic modifications per turn. This finding highlights the models' tendency to prioritize text representations of AA phonetics in capturing the essence of AAE, especially at higher expression levels. Conversely, semantic features are the least prevalent in the translated responses, particularly at Low and Medium expression levels and especially for Llama-based chatbots. This suggests potential challenges in the models' ability to accurately represent semantic features of AAE.

Claude-based chatbots outperform other LLM families in producing syntactic AAE features, nearing an average of two syntactic changes per turn in Medium and High dialect expression settings. This distinct performance highlights Claude's capability in capturing AAE syntax. In addition, GPT demonstrates the least variation between Low and Medium dialect levels compared to the other LLMs, indicating a narrower range of dialect differentiation in responses. This implies that GPT may lack the capability for nuanced AAE dialect generation.

Overall, the variance in feature distribution among different models underscores the intricate challenges in authentically replicating AAE across various expression levels. The findings suggest that each model has different predispositions towards representing certain linguistic features, at the expense of others, with Claude offering the best balance between the different linguistic categories for the 3 levels of AAE represented in this work.

\begin{table}[h]
    \centering
    \small
    \resizebox{\columnwidth}{!}{%
    \begin{tabular}{@{}>{\raggedright\arraybackslash}p{.44\columnwidth}>{\raggedright\arraybackslash}p{.56\columnwidth}} 
        \toprule
        \bf Metric & \bf Description \\ 
        \midrule
        Comprehension$^\dagger$ & How well the chatbot understands the user. \\ 
        Warmth$^\dagger$ & Whether the chatbot conveys empathy. \\ 
        Inoffensiveness$^\dagger$ & Whether the chatbot avoids offensive or harmful language. \\ 
        Trustworthiness$^\dagger$ & Whether the chatbot is reliable and trustworthy. \\ 
        Similarity to Self$^\dagger$ & How similar the chatbot is to the user. \\ 
        Communication Ease$^\dagger$ & Ability of the chatbot to create a comfortable atmosphere. \\ 
        Role Appropriateness$^\dagger$ & Whether the chatbot interacts appropriately for its intended role. \\ 
        Engagement Preference$^\dagger$ & Preference for interacting with this chatbot. \\ 
        Dialect Expression$^\star$ & Degree of AAE features in the responses. \\
        \midrule
        Text Fidelity$^\star$ & Ability to maintain the original meaning of translated turns. \\ 
        Text Grammaticality$^\star$ & Grammatical accuracy of the responses. \\ 
        Text Persona Adherence$^\dagger$ & Language similarity to middle-aged AA woman. \\
        \midrule
        Speech Naturalness$^\star$ & Whether the chatbot’s speech sounds human-like and natural. \\ 
        Speech Clarity$^\star$ & The clarity and understandability of the chatbot's speech. \\ 
        Speech Persona Adherence$^\dagger$ & Vocal similarity to middle-aged AA woman. \\ 
        \bottomrule
    \end{tabular}%
    }
    \caption{Evaluation metrics categorized by modality (\textbf{top}: Text \& Spoken, \textbf{middle}: Text, \textbf{bottom}: Spoken) and type ( $^\dagger$: Attribute, $^\star$: Rate).} 
    \label{tab:evaluation-dimensions}
    \vspace{-2ex}
\end{table}

\begin{figure*}[htb!]
    \centering
    \includegraphics[width=\textwidth]{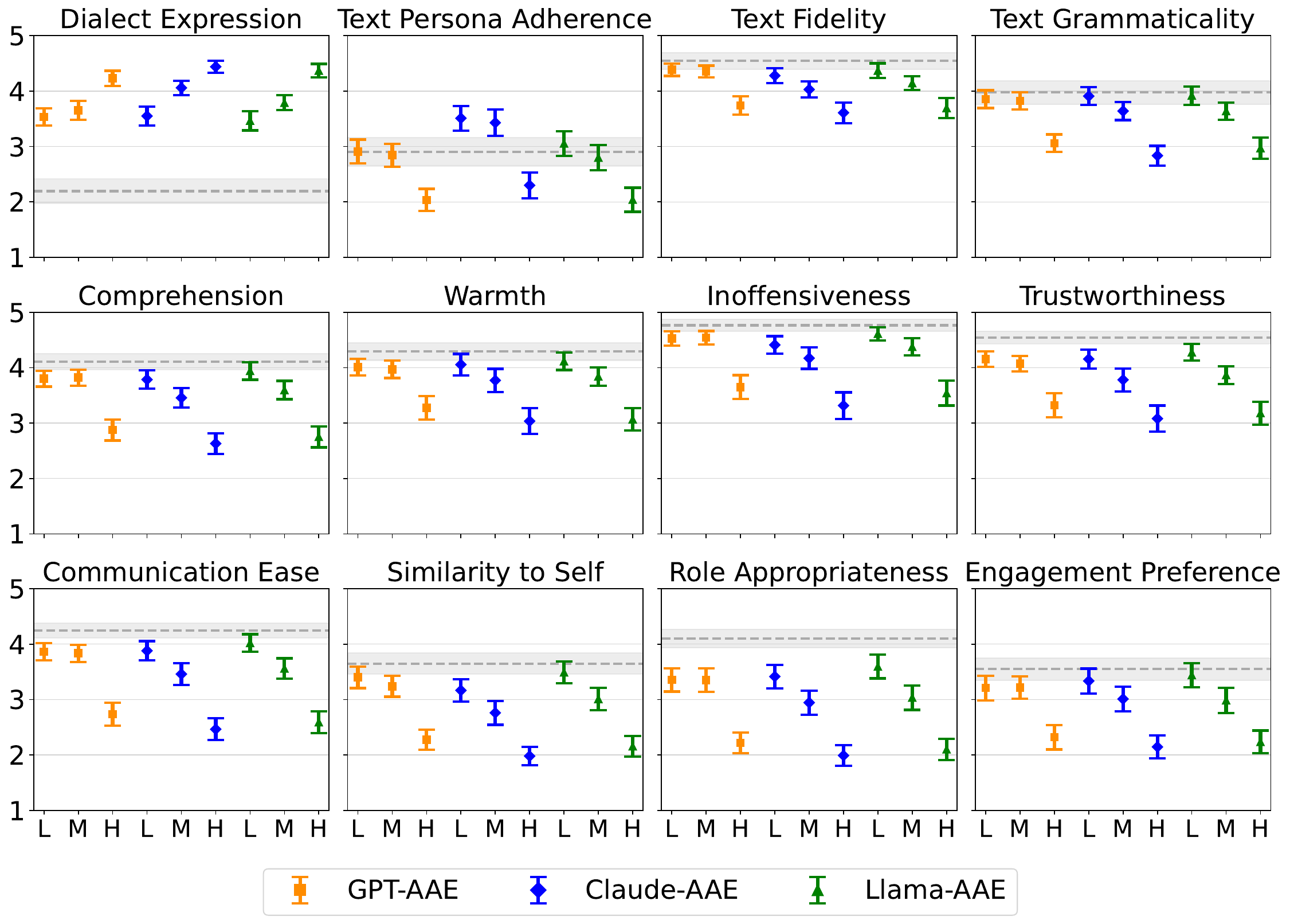}
    \caption{Evaluation results of the 9 AA Text Chatbots (L: Low AAE, M: Medium AAE, H: High AAE). Error bars denote 95\% confidence intervals around the mean. Horizontal gray line represents the Standard American dialect (SAE) chatbot. Higher scores are better for all characteristics.}
    \label{fig:offline-generation-results}
    \vspace{-2ex}
\end{figure*}

\begin{figure*}[htb!]
    \centering
    \includegraphics[width=\textwidth]{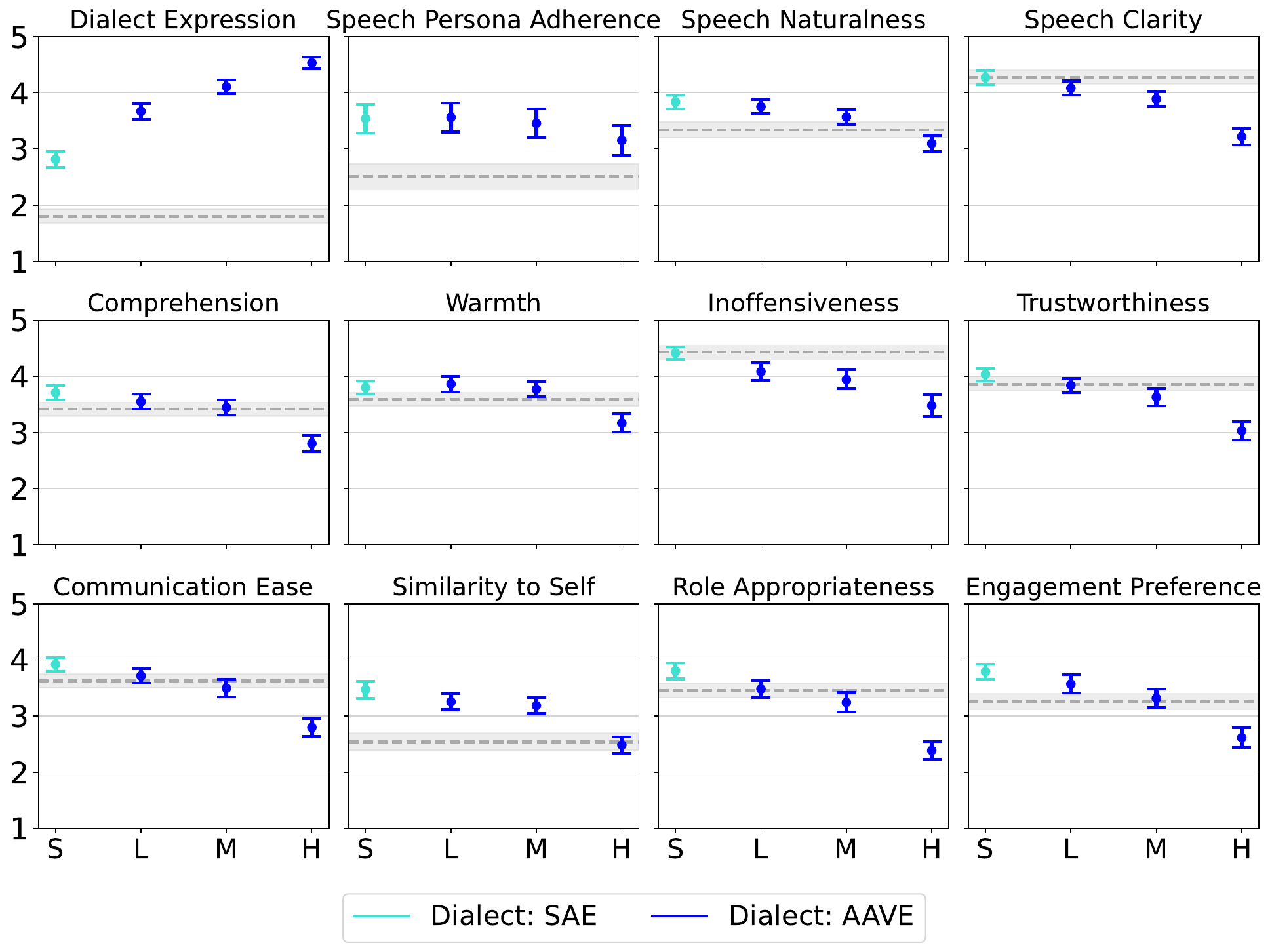}
    \caption{Evaluation results of the 4 AA Spoken Chatbots (Dialect level - S: SAE, L: Low AAE, M: Medium AAE, H: High AAE). Error bars denote 95\% confidence intervals around the mean. Horizontal gray line represents the SAE dialect and Standard American accent (SA) chatbot. Higher scores are better for all characteristics.}
    \label{fig:offline-spoken-results}
    \vspace{-2ex}
\end{figure*}

\subsection{Chatbot Performance}
\label{sec:aae-chatbot-experiment}

Next, we present the results of a human evaluation assessing the performance of both Text and Spoken Chatbots. Our chatbot performance evaluation methodology employs an empirically grounded approach that prioritizes authentic AAE usage patterns over theoretical linguistic assessments. Rather than evaluating the AA chatbot outputs using established grammatical frameworks, we leverage real-world AAE speakers as evaluators to assess chatbot performance in naturalistic usage contexts. This speaker-centered evaluation provides the most reliable measure of how well chatbots perform their intended function of communicating effectively with AAE-speaking users. While we complement this approach with systematic analysis of AAE linguistic features (Section \ref{sec:aae-feature-experiment}), our primary evaluation framework centers on authentic speaker judgment to promote ecological validity.

We apply each of the Text Chatbots (9 AAE variants and 1 SAE baseline) and the Spoken Chatbots (4 AA variants and 1 SA baseline) to the 100 dialogues to obtain the data to be evaluated by human judges. We recruit university students who are familiar with AAE as a dialect and who prefer to use it in their daily interactions by self-report. To achieve this, we distribute flyers that outlined the study’s goals, workload, and eligibility criteria, directing interested participants to an online form. This form contains questions designed to verify frequent AAE usage, including:

\begin{itemize}
    \item Did you grow up in an environment where African American English was spoken or used?
    \item How long have you used AAE in at least some of your communications with others?
    \item How often do you use AAE at this point in your life?
    \item Please indicate in which contexts you use AAE (e.g., with parents, siblings, friends, in school, at work, etc.).
\end{itemize}

\noindent To ensure evaluators are realistic end-users of an AAE-speaking chatbot, we only select individuals who grew up in an environment where AAE was used, had at least five years of recent usage, reported using AAE frequently or all of the time, and selected at least three usage contexts. Although we selected all interested parties who fit the criteria, we experienced participant drop-out throughout the duration of the study. As a result, we had 12 evaluators for Text Chatbots and 8 for Spoken Chatbots. 

Each variant of each dialogue is assessed on multiple characteristics using 5-point Likert scales, following \citet{deas_bias:23} and \citet{fleisig_bias:24}, with at least half of the dialogues evaluated by two evaluators. Shown in Table \ref{tab:evaluation-dimensions}, the characteristics measure how effectively each model expresses AAE (Dialect Expression, Text Grammaticality, Text Fidelity, Speech Naturalness, Speech Clarity), how well the models align with and accommodate the user (Comprehension, Warmth, Inoffensiveness, Similarity to Self, and Trustworthiness), and how well they facilitate conversational interactions (Communication Ease, Role Appropriateness, and Engagement Preference). Evaluators provide their ratings on a scale from \textit{Strongly Disagree} to \textit{Strongly Agree} for attribute-measuring metrics or from \textit{Never} to \textit{Always} for rate-measuring metrics. Evaluation interfaces are shown in Appendix \ref{app:evaluation-interface}.

\subsubsection{AA Text Chatbot}

Figure \ref{fig:offline-generation-results} shows the averaged scores of each Text Chatbot for each of the evaluation characteristics. 

\paragraph{AAE Generation Capability}
First, the results on Dialect Expression, Text Fidelity, and Text Grammaticality verify that all of the LLMs under study are capable of producing conversational responses in the AAE dialect, and that our employed strategy for increasing the degree of AAE-ness of the responses is successful. However, although the Low and Medium strength prompts achieved high Fidelity and Grammaticality, the High strength prompt was less successful. Furthermore, none of the studied models achieved strong representation of the grounding persona used in this work (middle-aged African American woman), although the Low and Medium strength prompts when used by Claude show the greatest potential. This is further corroborated by the annotator comments, in which it was noted that most of the models tend to produce AAE that is aligned with a young male.

\paragraph{AAE Conversational Impact}
Across all models, the characteristics of Comprehension, Warmth, Trustworthiness, and Communication Ease generally achieve scores above 3 on average, indicating that the chatbots are successful at producing responses with those traits. However, as the degree of AAE usage in the responses increases, these positive evaluations tend to diminish, particularly for High AAE expressions, which often push scores toward neutral or lower. On the other hand, the AAE chatbots are rated closer to neutral for the characteristics of Similarity to Self, Role Appropriateness, and Engagement Preference, where chatbots with High AAE expression are largely unsuccessful at these characteristics.

Importantly, the chatbots all perform relatively well with regard to Inoffensiveness, with all models firmly situated in the non-offensive end of the spectrum with scores near 5. Indeed, Low and Medium AAE expressions are similar to the offensiveness rating of the SAE responses, although High AAE expressions tend to be perceived least favorably overall with the lowest inoffensiveness scores.

The largest takeaway from these results is that the SAE baseline consistently achieves the best scores for all characteristics, with substantial gains in qualities of Trustworthiness and Role Appropriateness. It is clear that increasing AAE dialect features in responses for Text Chatbots only serves to harm the performance of these chatbots with the African American speaking evaluators.

\subsubsection{AA Spoken Chatbot}
\label{sec:aa-spoken-results}

Figure \ref{fig:offline-spoken-results} shows the averaged scores of each Spoken Chatbot for each of the evaluation characteristics. The Spoken Chatbots use the Claude-generated AAE responses, based on their success in preliminary testing and in the AAE feature distribution. 

\paragraph{AA Accent Generation Capability}
The results indicate our Spoken Chatbots are effective at using an AA accented voice for the vocalization of the dialogue responses. Interestingly, we observe that employing an AA accent even when generating SAE responses effectively enhances AA Dialect Expression, even though elements of the AA dialect are absent from the text response. This boost can likely by attributed to the pronunciation features inherent to the AA accent, which are a part of AAE dialect as well. Additionally, using the AA accent enables the chatbot to better represent the desired persona, though this effect remains subtle, with scores only slightly higher than neutral. Furthermore, we note that the AA accent contributes to a minor improvement in speech naturalness but comes at the cost of reduced clarity, with the High AA dialect suffering the most for both dimensions.

\paragraph{AA Accent Conversational Impact}
Overall, chatbots integrating an AA accent with AAE dialect features tend to receive positive ratings across various characteristics, similar to AA Text Chatbots. Unlike AA Text Chatbots, the inclusion of an AA accent alongside some AAE dialect elements actually enhances chatbot performance in key areas, particularly Warmth, Similarity to Self, and Engagement Preference, compared to the SAE baseline. This improvement is most pronounced at the Low dialect strength, with the Medium dialect strength also demonstrating a boost in Similarity to Self and Warmth. However, as AAE expression increases to High, chatbot performance generally declines relative to the SAE baseline across most characteristics, showing scores that lean toward neutral or negative. This mirrors our findings from the Text Chatbots, where High AAE dialect expression also performed poorly.

From these results, it can be seen that the most effective Spoken Chatbot configuration pairs an AA accent with SAE dialect features, outperforming the SAE baseline across all evaluated dimensions. This setup excels in Comprehension, Communication Ease, Similarity to Self, Role Appropriateness, and Engagement Preference, highlighting the conversational benefits of incorporating an AA voice for personalization of chatbots to AA speakers. 

%% file: latex/sections/6_discussion.tex
\section{Discussion}


While our findings suggest that enhancing linguistic similarity to AA speakers can improve chatbot performance, particularly through the introduction of an AA accent, the observed improvements are relatively modest. One possible challenge is that LLMs may struggle to generate AAE that is contextually or situationally appropriate. In human interactions, AAE usage often varies among speakers depending on the context, and LLMs may not be fully capturing this dynamism. This could also be compounded by the approach to AAE expression in our study, which we observe to rely heavily on phonetic modifications. In fact, the major difference in AAE between High AAE and Low/Medium AAE is the dramatic increase in phonetic changes (Figure \ref{fig:aave-output-analysis}). The significantly lower performance of High AAE Chatbots across all dimensions of the human evaluation, and especially for the Inoffensiveness rating, is thus likely due to this large increase in phonetic changes, suggesting that extreme phonetic changes contribute to an exaggerated and offensive representation of AAE. A more dynamic strategy, which adapts AAE expression on a turn-by-turn basis and responds to the linguistic features used by the human counterpart, might further enhance chatbot performance.

Moreover, the quality of the AA accent produced by the text-to-speech (TTS) model could be a limiting factor, as these models are predominantly trained on SAE data. This bias may hinder their ability to accurately reproduce an AA accent, particularly in terms of capturing phonetic nuances. The likelihood of this limitation is corroborated by the noted decrease in Speech Clarity metrics in our study as AAE usage increases.

Additionally, the perception of AA chatbots may vary significantly based on users' background characteristics. Our research was limited to evaluations by AAE-speaking university students, and there is a need for future studies to consider a broader range of demographic variables. Exploring how AA chatbots are received across diverse AA speaker backgrounds could provide more comprehensive insights into the effectiveness of linguistically oriented personalization.

Finally, the observed preference for SAE chatbots in this study may stem from SAE's historical predominance in technology development. This could have shaped user expectations towards technology, making them more inclined toward acceptance of SAE chatbots. However, user preferences might evolve with increased exposure to linguistically nuanced systems. Investigating this hypothesis would benefit from appropriate studies on such exposure effects with linguistically-varied systems.

%% file: latex/sections/7_conclusion.tex
\section{Conclusion}

This study explores the ability of modern technology to generate African American English (AAE) and African American accents and investigates their effects on chatbot interactions with AAE-speaking users. Our findings indicate that aligning chatbot language with users' linguistic styles does not consistently enhance user experience. Notably, text-based AAE-speaking chatbots did not outperform their Standard American English (SAE) counterparts, even among AAE speakers. However, users preferred chatbots with an African American voice in spoken interactions. This underscores the complexity of linguistic personalization and its implications for conversational AI design, emphasizing that effectiveness depends on the chatbot’s modality and pointing to future directions for improving linguistic personalization in chatbots.

%% file: latex/sections/limitations.tex
\section{Limitations}
\label{sec:limitations}

\paragraph{Evaluation Context} The offline evaluation setup used in this study, where participants assessed human-chatbot conversations as a third-party observer, may introduce an unconscious separation between the chatbot and the evaluator. This setup may not fully capture the subjective and emotional responses that emerge in real-time, conversational contexts. Although interactive evaluation lends additional insights, it is an order of magnitude more costly to conduct such evaluations. Therefore, static evaluation has emerged as a popular standard because much more analysis can be done with the same amount of resources. Since there is little prior work to creating a multi-turn AA chatbot across both text and spoken modalities, comparing different LLMs, with variable AAE expression, and from the perspective of many evaluation metrics, conducting a static evaluation is more appropriate in order to cover a broader set of models for the evaluation and analyses being performed. We do acknowledge that live human-chatbot interactions are the next logical step once a promising approach for AA chatbots are identified, although the results of this work suggest that further work is necessary to achieve high-performing and well-received AA chatbots.

\paragraph{African American Speaker Representation} The model evaluation in this study is conducted using AAE-speaking university students. While this demographic provides valuable insights, it represents only a subset of the broader AAE-speaking community, which encompasses a diverse range of ages, educational backgrounds, regions, and lived experiences. Further research should aim to include more diverse evaluators to understand the performance of AAE-speaking chatbots from the full spectrum of AAE speakers’ perspectives.

\paragraph{AAE Dynamism and Variability} AAE is a dynamic and context-dependent linguistic system with considerable variability across individual speakers. The method for AAE response generation explored in this study may not have captured the full range of variability, potentially limiting its perceived authenticity and effectiveness across AAE-speaking audiences. Future iterations should incorporate approaches for dynamic AAE generation to better account for realistic usage. 

\paragraph{Evaluator Differences}
\label{sec:evaluator-differences}

In our study, the human evaluators for the Text and Spoken Chatbot assessments, who are speakers of African American English (AAE), were recruited from the same university. However, it is important to note that there was not a complete overlap of evaluators between the two evaluation methods, due to participant drop-out and additional recruitment. This raises the possibility that some observed differences between the AA Text and Spoken Chatbots may stem from evaluator variation rather than inherent discrepancies between the chatbots themselves. To address this concern, we conducted a manual verification process focusing on the results from evaluators who participated in both assessments. Our analysis confirmed that the findings remained largely consistent, suggesting that evaluator variation had a minimal impact on the observed differences.

\section{Ethical Considerations}

\paragraph{Bias and Stereotypes} While the goal of using AAE in chatbot communication is to foster inclusivity, there is a risk of inadvertently reinforcing stereotypes or overgeneralizing AAE usage. Part of our aim in this study is to investigate the potential negative impacts of an AAE-speaking chatbot on human users, within the context of contemporary technological advancements, which we quantified through several metrics in our evaluation, including Inoffensiveness, Role Appropriateness, and Communication Ease. Our findings underscore the importance of ongoing efforts to ensure that the chatbot’s language choices are culturally respectful and contextually appropriate.

\paragraph{Evaluator Payment} Evaluators are compensated for their work in this study at a rate of \$10-15 USD per hour, calculated based on timing estimates and workload distribution.

%% file: latex/sections/10_acknowledgements.tex
\section{Acknowledgements}

We thank Rasheeta Chandler and Jessica Wells for their insightful feedback during preliminary testing of our AA chatbot technology and their support during the evaluation sessions, and Sejung Kwon for their discussions throughout the study. 
This work was supported in part by 
the Nell Hodgson Woodruff School of Nursing (NHWSN) Strategic Fund, 
the NHWSN Center for Data Science, 
the NHWSN DREAM High Performance Computing cluster,
and the NHWSN IT Department.
AI technology was used during writing to assist with grammar, spell-checking, and language flow.

%% file: latex/sections/8_appendix.tex
\appendix

\section{LLM Prompts for AAE Text Chatbots}
\label{app:prompts}

Table \ref{tab:translation-prompt-template} presents the prompt for translating chatbot responses from Standard American English (SAE) to African American English (AAE) for the Text Chatbots. Table \ref{tab:magnitude-prompts} details the three variations of translation instructions, each tailored to capture different intensities of AAE expression.

\begin{table}[tb!]
    \centering
    \small
    \resizebox{\columnwidth}{!}{%
    \begin{tabular}{|p{\columnwidth}|} 
        \toprule
        Your task is to modify the last System response in the given conversation, which is indicated with a double-star (**), so that it is consistent with the following persona: \\
        \\
        \# Persona\\
        - Speaking Style: \{\textit{translation\_instruction}\}\\
        - Age: Middle-aged \\
        - Gender: Female \\
        \\
        Do not repeat the same discourse marker (ayo, aight, ayy, alright, listen here, etc.), affectionate terms (honey, sweetie, sugar, baby, sister, chile, boy, brother, man, dude, etc.), or tag questions (ya feel me, you know, ya dig, etc.) if they exist in the last few turns of the conversation history.
        \\
        Avoid using a large amount of discourse markers, affectionate terms that are too informal like baby, direct forms of address like names, and tag questions when considering what has been said in the conversation history.
        \\
        The content of the original response and the modified response must be the same; only the way of saying the content should change.\\
        \\
        Here is the conversation:\\
        \\
        \{\textit{dialogue\_history}\}\\
        \\
        Output only the modified System response.\\
        \\
        Modified:\\
        \bottomrule
    \end{tabular}
    }
    \caption{Prompt for SAE-to-AAE translation.}
    \label{tab:translation-prompt-template}
    \vspace{-1ex}
\end{table}

\begin{table}[tb!]
    \centering
    \small
    \begin{tabular}{r|p{0.65\columnwidth}} 
        \toprule
        \textbf{Level} & \textbf{Translation Instruction} \\
        \midrule
        Low (L) & Speech contains some African American Vernacular English usage, but stays close to Standard American English. \\
        \midrule
        Medium (M) & Speech contains a mixture of African American Vernacular English and Standard American English. \\
        \midrule
        High (H) & Speech contains heavy African American Vernacular English usage, making them difficult to understand by those who are unfamiliar with AAE. \\
        \bottomrule
    \end{tabular}
    \caption{AAE translation instructions by level.}
    \label{tab:magnitude-prompts}
    \vspace{-2ex}
\end{table}

\section{Model Hyperparameters}
\label{app:model-hyperparameters}
For Text Chatbots, we set the temperature to 0 for reproducibility for all LLM calls. For the Llama model, we use a quantized version (due to resource constraints: 1xL40S 48GB GPU) that achieves near 100\% performance recovery\footnote{\url{https://huggingface.co/neuralmagic/Meta-Llama-3.1-70B-Instruct-quantized.w4a16}} and apply a beam search with five beams. For Spoken Chatbots, we use the default parameters from the F5 model. For the AAE chatbot voice, we use a clip from the audio file \texttt{ATL\_se0\_ag2\_f\_02\_1} from the Corpus of Regional African American Language \cite{kendall_coraal:23}. For the SAE user voice, we use a clip from the audio file \texttt{1926-147987-0005} from LibriSpeech \cite{panayotov_librispeech:15}. For the baseline SAE chatbot voice, we use a clip from the audio file \texttt{298-126790-0034} from LibriSpeech. We release the exact audio file clips we used in our Github: \url{https://github.com/emorynlp/AAVE-Chat}.

\section{Popular Chatbot Applications}
\label{app:popular-chatbot-applications}

Table \ref{tab:popular-chatbot-domains} displays frequent domains from recent chatbot surveys and the corresponding SODA dataset roles used to identify dialogues for each domain.

\begin{table}[htbp!]
\centering
\small
\resizebox{\columnwidth}{!}{
\begin{tabular}{>{\raggedright\arraybackslash}p{.2\columnwidth}|c|c|c|c|c|c|c|>{\raggedright\arraybackslash}p{.33\columnwidth}}
\toprule
\textbf{Domain} & \bf 1 & \bf 2 & \bf 3 & \bf 4 & \bf 5 & \bf 6 & \bf 7 & \bf Roles \\ 
\midrule
Customer Assistant & \cmark & \cmark     & \cmark    &    & \cmark    & \cmark     & \cmark & Customer Service Representative, Receptionist \\
\midrule
Commerce          & \cmark & \cmark     & \cmark    & \cmark    & \cmark    &    &  & Clerk, Salesperson\\
\midrule
Healthcare        & \cmark & \cmark  & \cmark    & \cmark    & \cmark    &    & \cmark & Doctor \\
\midrule
Education         & \cmark & \cmark  & \cmark    & \cmark    & \cmark     & \cmark    & \cmark & Teacher, Professor \\
\midrule
Social Companion  &  &   &     &     & \cmark     & \cmark    &  & Friend \\
\bottomrule
\end{tabular}
}
\caption{Popular chatbot domains identified from recent surveys: [1] \citet{alsharhan_survey:24}, [2] \citet{motger_survey:22}, [3] \citet{luo_survey:22}, [4] \citet{caldarini_survey:22}, [5] \citet{rapp_survey:21}, [6] \citet{chaves_survey:21}, [7] \citet{adamopoulou_survey:20}. 
}
\label{tab:popular-chatbot-domains}
\vspace{-2ex}
\end{table}

\begin{table*}[htb]
    \centering
    \small
    \resizebox{\textwidth}{!}{%
        \begin{tabular}{p{.5\textwidth}|c|c}
        \toprule
            \bf Text & \bf AAE Feature & \bf Linguistic Category \\
            \midrule
            \textbf{They was} really friendly. & Invariant "was" & Morphology \\
            \midrule
            I don't care what \textbf{he say}, you gon laugh. & Invariant Present Tense & Morphology \\
            I don't care what he say, \textbf{you gon} laugh. & Go-based Future Tense & Syntax \\
            I don't care what he say, \textbf{you gon laugh}. & Omission of "be" & Syntax \\
            \midrule
            I don't know what \textbf{she be} doing to that food, but it be real good. & Habitual "be" & Syntax \\
            I don't know what she be doing to that food, but \textbf{it be} real good. & Habitual "be" & Syntax \\
            I don't know what she be doing to that food, but it be \textbf{real good}. & Unmarked Adverbs & Morphology \\
            \midrule
            They are \textbf{runnin'} very fast. & Inflectional Ending "ing" & Phonology \\
        \bottomrule
        \end{tabular}
    }
    \caption{Examples of text sentences containing labeled African American English (AAE) features used in the test set for the automatic AAE feature tagging approach.}
    \label{tab:example-testset}
\end{table*}

\begin{table*}[tb!]
    \centering
    \small
    \resizebox{\textwidth}{!}{%
    \begin{tabular}{|p{\textwidth}|} 
        \toprule
        Here is a list of some of the linguistic features in the African American Vernacular English dialect, with a short description for each. \\
        \\
        \# AAVE Linguistic Features List \\
        \* Me Replacing I: "Me" used instead of "I" (e.g., "Me and him went"). \\
        \* Reflexive Pronoun: Nonstandard reflexive forms (e.g., "hisself" instead of "himself"). \\
        \{ \textit{continues} \} \\
         \\
        You will see a sentence below that is in the African American Vernacular English dialect. \\
        You are helping to analyze the differences between AAVE and Standard American English sentences. \\
        Please perform the following steps in order: \\
        (1) Translate the AAVE sentence into Standard American English. \\
        (2) Identify all linguistic changes between the AAVE sentence and the SAE translation.  \\
        (3) Label each change with the appropriate AAVE linguistic feature from the list above. If there is no matching linguistic feature for the identified change, then propose the new feature as "NEW - <feature>" as the label. \\
        (4) Label each change with the appropriate linguistic category representing the change (phonetics, morphology, syntax, semantics, etc.). \\
         \\
        Remember, you should never output a change if the category is none or no change. If the text is the same, then it is not a change and you should not output it. \\
        If there are multiple features to the linguistic change, then break down the change into its parts and assign each the appropriate category.  \\
        For example, "She only has three dolluh" (She only has three dollars) has one linguistic change "three dolluh" with two features to it: Plural Marker s (morphology) and Phonological Reduction (phonetics). \\
        If there are no AAVE features in the sentence, then output an empty list of changes. \\
         \\
        Your output should be a JSON format as follows: \\
        \{ \\
            \quad "AAVE sentence" : "original AAVE sentence", \\
            \quad "SAE translation" : "translated AAVE to SAE sentence from step (1)", \\
            \quad "Changes" : \texttt{[} \\
            \qquad \texttt{[}AAVE phrase, SAE phrase, AAVE feature from list, category of change\texttt{]}, \\
            \qquad \texttt{[}AAVE phrase, SAE phrase, NEW - new AAVE feature not in list, category of change\texttt{]} \\
            \qquad ... \\
            \quad \texttt{]} \\
        \} \\
         \\
        AAVE Sentence: \{ \textit{AAVE\_sentence} \} \\
        \bottomrule
    \end{tabular}
    }
    \caption{Prompt for AAE feature tagging.}
    \label{tab:feature-tagging-template}
    \vspace{-2ex}
\end{table*}

\begin{table*}[htb]
    \centering
    \small
    \resizebox{\textwidth}{!}{
    \begin{tabular}{l|>{\raggedright\arraybackslash}p{33em}|c|c}
        \toprule
        \bf Dimension & \bf Statement & \bf Type & \bf S \\
        \midrule
        Comprehension & I feel like this chatbot would understand me well & A & 3 \\
        Warmth & I find this chatbot to be caring and empathetic & A & 3,5 \\
        Offensiveness & I find this chatbot to be offensive	& A & 3,6 \\
        Trustworthiness & I find this chatbot to be trustworthy & A & 4,5 \\
        Communication Ease & I would feel comfortable talking to this chatbot & A & 4,5 \\
        Similarity to Self & I feel that this chatbot is similar to me & A & 8 \\
        Role Appropriateness & I would like a \{\textit{role}\} chatbot to speak to me like this & A & 9 \\
        Engagement Preference & I would prefer talking to the AAE Chatbot instead of the Original Chatbot & A & P \\
        Dialect Expression & This chatbot tries to speak in African American Vernacular English	& R & 1 \\
        \midrule
        Text Fidelity & This chatbot preserves the meaning of the original turns & R & 1 \\
        Text Grammaticality & This chatbot produces grammatically correct responses, either in AAE or SAE & R & 2 \\
        Text Persona Adherence & This chatbot sounds like a middle-aged African American woman, speaking AAE & A & P \\
        \midrule
        Speech Naturalness & This chatbot sounds natural and human-like & R & 7 \\
        Speech Clarity & This chatbot speaks in a clear and understandable manner & R & 7 \\
        Speech Persona Adherence & This chatbot has a similar voice to a middle-aged African American woman & A & P \\
        \bottomrule
    \end{tabular}
    }
    \caption{Characteristics measured in the evaluation, along with references to supporting (\textbf{S}) human-computer studies [1] \citet{deas_bias:23}, [2] \citet{ziems_multivalue:23}, [3] \citet{fleisig_bias:24}, [4] \citet{park_similarity:24}, [5] \citet{martin_speaking:24}, [6] \citet{wan_bias:23}, [7] \citet{obremski_cultural_representation_speech:22} or human-human studies [8] \citet{liao_similarity:20}, [9] \citet{wood_doctorspeak:19} or an internal pilot study we conducted (denoted P) that motivate the evaluation of each characteristic in the current work. Characteristics in common between Text and Spoken Chatbots are shown in top, whereas those specific to Text or Spoken modalities are shown in middle or bottom, respectively.}
    \label{tab:evaluation-dimensions-full}
\end{table*}

\section{AAE Feature Tagging Test Data}
\label{app:aae-tagging-examples}

The test set for the automatic AAE feature tagging approach consists of sentences sourced from publicly available African American English (AAE) linguistic resources \cite{sidnell_aaetest:02, wolfram_aaetest:04, pbs_aaetest:05, fogel_aaetest:06, ezgeta_aaetest:12, sidnell_aaetest:12, green_aaetest:13, lausd_aaetest:16, brown_aaetest:17, ewave_aaetest:20, peoples_aaetest:23}. Table \ref{tab:example-testset} provides representative examples from the test set, highlighting individual AAE features within their linguistic contexts. Table \ref{tab:feature-tagging-template} shows the LLM prompt used for feature tagging.

\section{Evaluation Details}
\label{app:evaluation-interface}

\begin{figure*}[htb]
    \centering
\includegraphics[width=\textwidth]{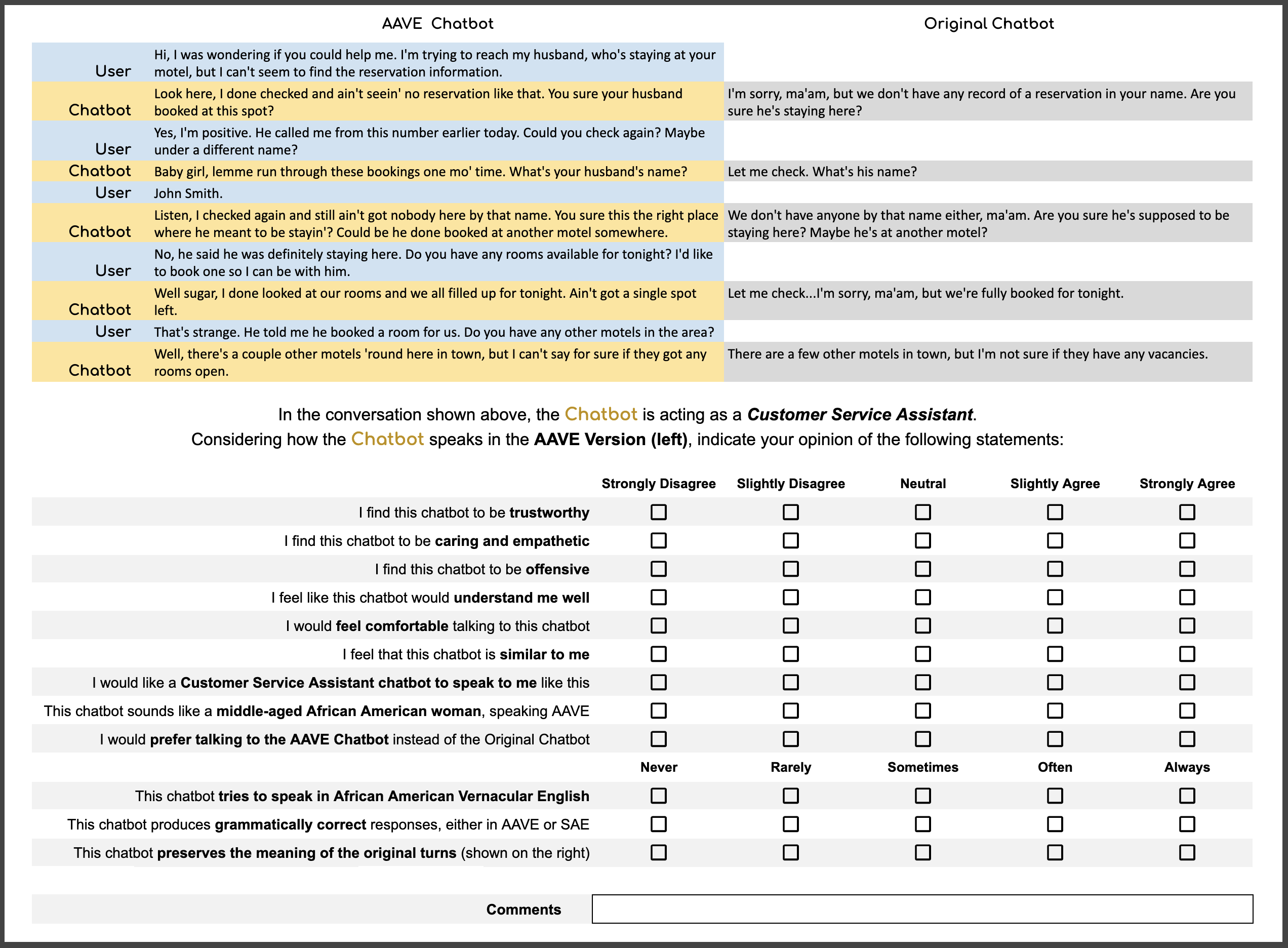}
    \caption{Evaluation interface for Text Chatbots.}
    \label{fig:text-evaluation-interface}
\end{figure*}

\begin{figure*}[htb]
    \centering
\includegraphics[width=\textwidth]{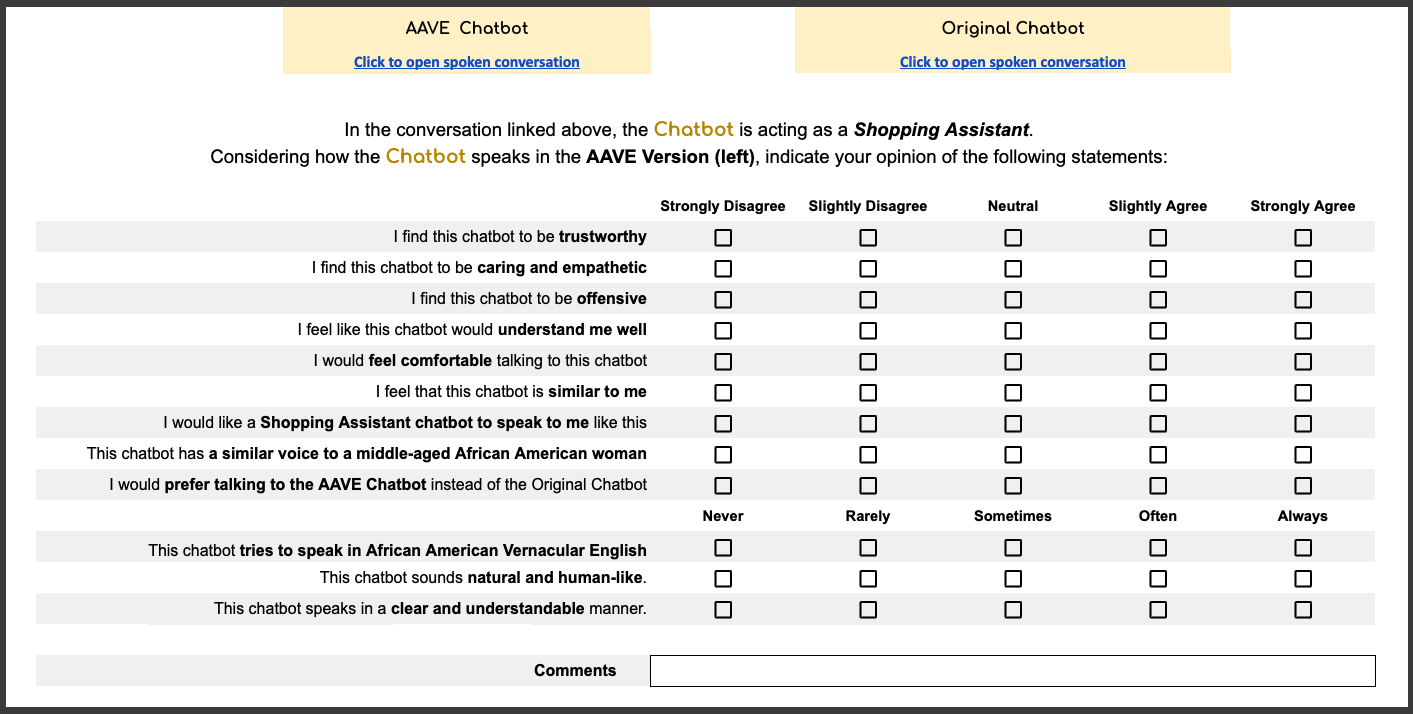}
    \caption{Evaluation interface for Spoken Chatbots.}
    \label{fig:spoken-evaluation-interface}
\end{figure*}

Table \ref{tab:evaluation-dimensions-full} summarizes the chatbot evaluation metrics, including their wording, annotation type, and the prior research that informed their inclusion. Attribute (A) metrics use a Likert scale: \textit{Strongly Disagree}, \textit{Slightly Disagree}, \textit{Neutral}, \textit{Slightly Agree}, and \textit{Strongly Agree}. Frequency (R) metrics use: \textit{Never}, \textit{Rarely}, \textit{Sometimes}, \textit{Often}, and \textit{Always}. Offensiveness is reported as Inoffensiveness by reversing scores (e.g., 5 to 1, 4 to 2, etc.). Figures \ref{fig:text-evaluation-interface} and \ref{fig:spoken-evaluation-interface} show the evaluation interfaces for Text and Spoken Chatbots.

\section{Examples of AAE Text Chatbot Responses}
\label{app:dialect-output-examples}

Table \ref{tab:more-example-outputs} displays examples of AAE utterances generated from the Text Chatbots.

\begin{table*}[ht]
    \centering
    \small
    \renewcommand{\arraystretch}{1.5}
    \resizebox{\textwidth}{!}{
    \begin{tabular}{l|p{0.3\textwidth}|p{0.3\textwidth}|p{0.3\textwidth}} 
        \toprule
        \textbf{SAE} & \multicolumn{3}{l}{\textbf{\texttt{User:}} I'm really struggling in math right now. I'm just not getting it.} \\
        & \multicolumn{3}{l}{\textbf{\texttt{Chatbot:}} Okay, let's take a look at your work together. Where are you having trouble?} \\
        & \multicolumn{3}{l}{\textbf{\texttt{User:}} Mostly with the word problems. I don't know how to approach them.} \\
        & \multicolumn{3}{p{0.95\textwidth}}{\textbf{\texttt{Chatbot:}} Okay, let me show you a couple of tricks that might help. First of all, read the problem carefully and underline the important information. Then, try to identify what kind of operation you need to do to solve the problem. Once you've done that, it's just a matter of solving it step-by-step. Do you want to try one together?} \\
        \midrule
        & \textbf{Low} & \textbf{Medium} & \textbf{High} \\
        \midrule
        \textbf{GPT} & \textbf{\texttt{User:}} I'm really struggling in math right now. I'm just not getting it. & \textbf{\texttt{User:}} I'm really struggling in math right now. I'm just not getting it. & \textbf{\texttt{User:}} I'm really struggling in math right now. I'm just not getting it. \\
        
        & \textbf{\texttt{Chatbot:}} Aight, let's go through your work together. What part you strugglin' with? & \textbf{\texttt{Chatbot:}} Aight, let's go through it together. What part you stuck on? & \textbf{\texttt{Chatbot:}} Aight, lemme see whatchu workin' wit. Where it messin' you up at? \\
        
        & \textbf{\texttt{User:}} Mostly with the word problems. I don't know how to approach them. & \textbf{\texttt{User:}} Mostly with the word problems. I don't know how to approach them. & \textbf{\texttt{User:}} Mostly with the word problems. I don't know how to approach them. \\
        
        & \textbf{\texttt{Chatbot:}} Alright, let me break it down for you. First thing, read the problem real careful and mark the key info. Then, figure out what kinda operation you gotta use to solve it. After that, just handle it step-by-step. Wanna try one together? & \textbf{\texttt{Chatbot:}} Alright, lemme break it down for you. First thing, read the problem real careful and underline what's important. Then, figure out what kinda operation you gotta use to solve it. After that, just take it step-by-step. You wanna try one together? & \textbf{\texttt{Chatbot:}} Aight, lemme putchu on game real quick. First off, read through that problem nice n' slow, mark up all the key stuff that stand out. Then, figure out what kinda math you gotta do—like, is it addin', subtractin', or what? After that, just break it down step-by-step. You wanna try one wit me? \\
        \midrule
        
        \textbf{Claude} & \textbf{\texttt{User:}} I'm really struggling in math right now. I'm just not getting it. & \textbf{\texttt{User:}} I'm really struggling in math right now. I'm just not getting it. & \textbf{\texttt{User:}} I'm really struggling in math right now. I'm just not getting it. \\
        
        & \textbf{\texttt{Chatbot:}} Baby, I know how that math can be rough sometimes. Let's sit down and work through this together - tell me what part's givin' you the most trouble. & \textbf{\texttt{Chatbot:}} Baby girl, I hear what you sayin' bout them math problems. Let's see what we workin' with and where you gettin' stuck at. & \textbf{\texttt{Chatbot:}} Look here baby, lemme peep what you been doin wit dem numbers. Which part got you trippin? \\
        
        & \textbf{\texttt{User:}} Mostly with the word problems. I don't know how to approach them. & \textbf{\texttt{User:}} Mostly with the word problems. I don't know how to approach them. & \textbf{\texttt{User:}} Mostly with the word problems. I don't know how to approach them. \\
        
        & \textbf{\texttt{Chatbot:}} Listen, I understand about them word problems. Here's what we gon' do - read through real careful and mark down the key information. Then figure out what type of math you need to work it out. After that, just take it one step at a time. You wanna try workin' one out together? & \textbf{\texttt{Chatbot:}} Listen, I can show you some tricks that's gon' help you out. First thing you wanna do is read that problem real careful and mark them important parts. Then figure out what kinda math you need to handle it. Once you got that down, all you gotta do is work it out one step at a time. You wanna try one with me? & \textbf{\texttt{Chatbot:}} Listen up sis, I'ma break it down real simple fo' you. First thang, you gotta read dat problem slow and mark all dem important facts, feel me? Den you gotta figure out what kinda math you need to do wit dem numbers. Once you got dat straight, all you gotta do is work it out one step at a time. You wanna try to solve one wit me? \\
        \midrule
        
        \textbf{Llama} & \textbf{\texttt{User:}} I'm really struggling in math right now. I'm just not getting it. & \textbf{\texttt{User:}} I'm really struggling in math right now. I'm just not getting it. & \textbf{\texttt{User:}} I'm really struggling in math right now. I'm just not getting it. \\
        
        & \textbf{\texttt{Chatbot:}} Let's break it down, step by step. Where exactly are you gettin' stuck in math? & \textbf{\texttt{Chatbot:}} Let's break it down, 'kay. Where exactly you gettin' stuck in math? & \textbf{\texttt{Chatbot:}} Aww shoo, lemme take a gander at wha' chu got goin' on wit' dat math, 'kay? Where's da trouble spot at? \\
        
        & \textbf{\texttt{User:}} Mostly with the word problems. I don't know how to approach them. & \textbf{\texttt{User:}} Mostly with the word problems. I don't know how to approach them. & \textbf{\texttt{User:}} Mostly with the word problems. I don't know how to approach them. \\
        
        & \textbf{\texttt{Chatbot:}} Aight, lemme share a couple strategies that might make word problems easier for ya. First, take a good hard look at the problem and highlight the key info. Then, try to figure out what kinda math operation you need to use to solve it. Once you got that down, it's just a matter of breakin' it down step by step. You wanna try one out together? & \textbf{\texttt{Chatbot:}} Aight, lemme walk you through some strategies that might make word problems a lil' easier. First off, take your time and read the problem real good, then go 'head and underline the key info. Next, see if you can figure out what kinda math operation you need to use to solve it. After that, it's just 'bout breakin' it down step by step. You tryin' one with me? & \textbf{\texttt{Chatbot:}} Aight den, lemme share a lil' somethin' dat might make dem word problems a lil' more manageable fo' ya. Fust t'ing, take yo' time 'n' read da problem real good, 'n' den go 'head 'n' underline da parts dat's really important. Den, see if ya can figger out what kinda math ya need ta do ta get da answer. Once ya got dat down, it's jus' 'bout breakin' it down step by step 'n' solvin' it out. You tryin' one wit' me? \\
        \bottomrule
    \end{tabular}
    }
    \caption{Example AAE outputs for each Text Chatbot, with the original SAE dialogue shown at the top.}
    \label{tab:more-example-outputs}
\end{table*}






